\documentclass[letterpaper, 10pt, conference]{ieeeconf}  

\IEEEoverridecommandlockouts                              

\usepackage[dvipsnames]{xcolor}
\usepackage{amsmath}
\usepackage{amssymb}
\usepackage{bm}
\usepackage{booktabs}
\usepackage{caption}
\usepackage{csquotes}
\usepackage{epsfig}
\usepackage{graphicx}
\usepackage{multirow}
\usepackage{tabularx}
\usepackage{times}
\usepackage{xspace}

\usepackage[pagebackref=true,breaklinks=true,letterpaper=true,colorlinks,bookmarks=false]{hyperref}
\usepackage[capitalize]{cleveref}

\newtoggle{nographics}
\newtoggle{lrgraphics}
\newtoggle{nocomments}
\newtoggle{supplementary}

\togglefalse{nographics}
\togglefalse{lrgraphics}
\togglefalse{nocomments}
\togglefalse{supplementary}

\crefname{section}{Section}{Sections}
\crefname{table}{Table}{Tables}
\crefname{figure}{Figure}{Figures}

\newcommand{\Kill}[1]{}
\iftoggle{nocomments}{
    \newcommand{\LAURENZ}[1]{\ignorespaces}
    \newcommand{\CEDRIC}[1]{\ignorespaces}
    \newcommand{\ALEXEY}[1]{\ignorespaces}
    \newcommand{\HAIFAN}[1]{\ignorespaces}
    \newcommand{\AK}[1]{\ignorespaces}
    \newcommand{\MATTHIAS}[1]{\ignorespaces}
    \newcommand{\todo}[1]{\ignorespaces}
    \newcommand{\plan}[1]{\ignorespaces}
}{
    \newcommand{\LAURENZ}[1]{\textbf{\textcolor{violet}{Laurenz: #1}}}
    \newcommand{\CEDRIC}[1]{\textbf{\textcolor{purple}{Cedric: #1}}}
    \newcommand{\ALEXEY}[1]{\textbf{\textcolor{orange}{Alexey: #1}}}
    \newcommand{\HAIFAN}[1]{\textbf{\textcolor{cyan}{Haifan: #1}}}
    \newcommand{\AK}[1]{\textbf{\textcolor{ProcessBlue}{AK: #1}}}
    \newcommand{\MATTHIAS}[1]{\textbf{\textcolor{red}{Matthias: #1}}}
    \newcommand{\todo}[1]{\textbf{\textcolor{red}{#1}}}
    \newcommand{\plan}[1]{\textsl{\textcolor{gray}{Say: #1}}}
}

\DeclareGraphicsExtensions{.eps,.pdf,.jpg,.png}
\iftoggle{lrgraphics}{
    \graphicspath{{src/media/lr/}{src/media/vector/}{src/media/tables/}}
}{
    \graphicspath{{src/media/hr/}{src/media/vector/}{src/media/tables/}}
}
\makeatletter
\iftoggle{nographics}{
    \LetLtxMacro{\includegraphics@orig}{\includegraphics}
    \RenewDocumentCommand{\includegraphics}{ s O{} m }{%
        {\setlength{\fboxsep}{0pt}%
         \colorbox{lightgray}{\phantom{\IfBooleanTF{#1}{\includegraphics@orig*}{\includegraphics@orig}[#2]{#3}}}%
        }%
    }
}{}
\makeatother

\crefname{figure}{Fig.}{Figs.}
\Crefname{figure}{Figure}{Figures}
\crefname{section}{Sec.}{Secs.}
\Crefname{section}{Section}{Sections}
\Crefname{table}{Table}{Tables}
\crefname{table}{Tab.}{Tabs.}

\makeatletter
\DeclareRobustCommand\onedot{\futurelet\@let@token\@onedot}
\def\@onedot{\ifx\@let@token.\else.\null\fi\xspace}

\def\eg{\emph{e.g}\onedot} 
\def\ie{\emph{i.e}\onedot}

\def\wrt{w.r.t\onedot} 

\makeatother

\NewDocumentCommand\rmse{}{%
    \ifmmode \text{RMSE}
    \else RMSE\xspace 
    \fi %
}
\NewDocumentCommand\qrmse{}{%
    \ifmmode \text{RMSE-}q_{95} 
    \else RMSE-$q_{95}$\xspace 
    \fi %
}
\NewDocumentCommand\fpr{}{%
    \ifmmode \text{FPR} 
    \else FPR\xspace 
    \fi %
}
\NewDocumentCommand\recall{}{%
    \ifmmode \text{Recall} 
    \else Recall\xspace 
    \fi %
}
\NewDocumentCommand\diracc{}{%
    \ifmmode \text{AngleAcc}
    \else AngleAcc\xspace 
    \fi %
}

\NewDocumentCommand\rmsev{}{%
    \ifmmode \text{RMSE}_{\text{v}}$
    \else RMSE$_{\text{v}}$\xspace 
    \fi %
}
\NewDocumentCommand\lpipsv{}{%
    \ifmmode \text{LPIPS}_{\text{v}}
    \else LPIPS$_{\text{v}}$\xspace 
    \fi %
}

\NewDocumentCommand\samplingdistance{}{%
    \ifmmode \lambda
    \else $\lambda$\xspace 
    \fi %
}

\begin{document}

\title{\LARGE \bf
AutoInst: Automatic Instance-Based Segmentation of LiDAR 3D Scans
}

\author{Cedric Perauer$^{1}$, Laurenz Adrian Heidrich$^{1}$, Haifan Zhang$^{1}$, \\
Matthias Nie{\ss}ner$^{1}$, Anastasiia Kornilova$^{2\text{*}}$, and Alexey Artemov$^{1\text{*}}$
\thanks{$^{1}$Cedric Perauer, Laurenz Heidrich, Haifan Zhang, Matthias Nie{\ss}ner, and Alexey Artemov are with the Technical University of Munich, Garching, Germany
{E-mail: artonson@yandex.ru}}%
\thanks{$^{2}$Anastasiia Kornilova is with the Skolkovo Institute of Science and Technology, Moscow, Russia.}%
\thanks{$^{\text{*}}$Equal senior author contribution.}%
\thanks{Anastasiia Kornilova was supported by the Analytical center under the RF Government (subsidy agreement 000000D730321P5Q0002, Grant No. 70-2021-00145 02.11.2021).}
}

\maketitle

\begin{abstract}
Recently, progress in acquisition equipment such as LiDAR sensors has enabled sensing increasingly spacious outdoor 3D environments.
Making sense of such 3D acquisitions requires fine-grained scene understanding, such as constructing instance-based 3D scene segmentations. 
Commonly, a neural network is trained for this task; however, this requires access to a large, densely annotated dataset, which is widely known to be challenging to obtain. 
To address this issue, in this work we propose to predict instance segmentations for 3D scenes in an unsupervised way, without relying on ground-truth annotations.
To this end, we construct a learning framework consisting of two components: (1) a pseudo-annotation scheme for generating initial unsupervised pseudo-labels;
and (2) a self-training algorithm for instance segmentation to fit robust, accurate instances from initial noisy proposals.
%
To enable generating 3D instance mask proposals, we construct a weighted proxy-graph by connecting 3D points with edges integrating multi-modal image- and point-based self-supervised features, and perform graph-cuts to isolate individual pseudo-instances. 
%
We then build on a state-of-the-art point-based architecture and train a 3D instance segmentation model, resulting in significant refinement of initial proposals. 
To scale to arbitrary complexity 3D scenes, we design our algorithm to operate on local 3D point chunks and construct a merging step to generate scene-level instance segmentations.
Experiments on the challenging SemanticKITTI benchmark demonstrate the potential of our approach, where it attains 13.3\% higher Average Precision and 9.1\% higher F1 score compared to the best-performing baseline. The code will be made publicly available at \newline \url{https://github.com/artonson/autoinst}.
\end{abstract}


\section{Introduction}
\label{sec:intro}

Operation of autonomous systems in real outdoor conditions involves continuously interpreting sensor data representing their 3D environment; commonly, an initial stage involves constructing a holistic 3D environment description by assigning semantic and instance labels to individual 3D points. Typically, neural networks (NNs)~\cite{marcuzzi2023mask,li2023cpseg} are trained for this task, requiring access to a large, densely annotated dataset, which is widely known to be costly and laborious to obtain~\cite{behley2019semantickitti, dai2017scannet, yeshwanthliu2023scannetpp}. 
Using neural models optimized on existing labeled collections~\cite{huang2023segment3d}, however, is suboptimal due to variations in sensor configurations and acquisition conditions across datasets, even when domain adaptation methods~\cite{bevsic2022unsupervised} are employed. 

\begin{figure}[t]
\centerline{
\includegraphics[
width=\columnwidth]{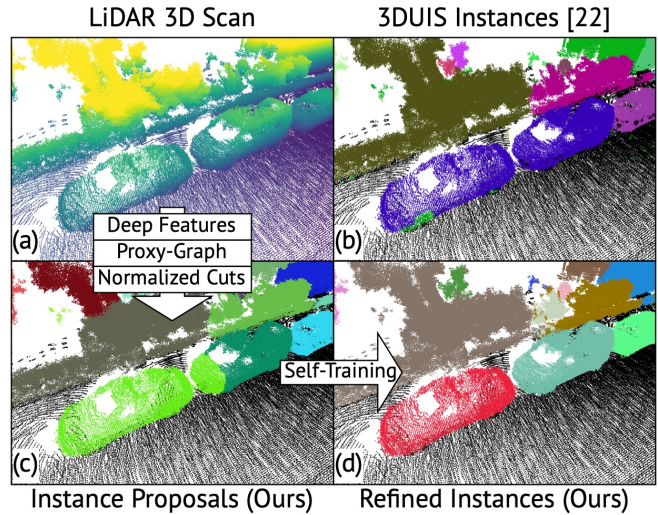}
}
\caption{For unsupervised instance segmentation of registered LiDAR 3D scans (a),
we integrate multi-modal self-supervised deep features into a weighted proxy-graph, making cuts for generation of instance mask proposals~(c) 
and performing their self-trained refinement~(d). 
Our algorithm is label-free and outperforms unsupervised baselines~(b).
\vspace{-1em}
}
\label{fig:teaser}
\end{figure}

\begin{figure*}[t]
\centering
\includegraphics[width=\textwidth]{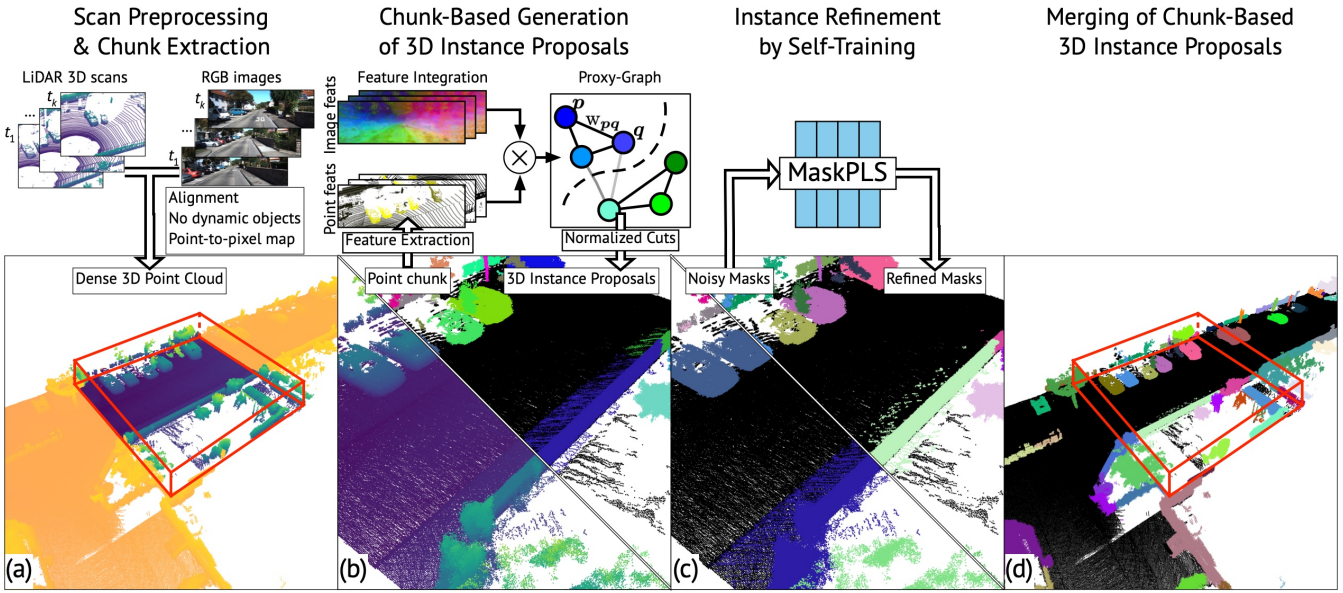}
\caption{
\textit{Overview of our unsupervised 3D instance segmentation framework.}
We start with a sequence of posed 3D LiDAR scans and RGB images, registering their static segments into a dense 3D map but operate with local overlapping chunks~(a).
To generate 3D instance mask proposals, we assign multi-modal features to individual 3D points, connecting them into a weighted proxy-graph; we cut the graph to obtain the coarse 3D mask proposals~(b).
For refinement of 3D instance masks, we start with the coarse proposals and perform multiple rounds of self-training, gradually reintegrating confident instance predicions as ground-truth~(c). 
Map-level segmentation is obtained by merging instances predicted in individual chunks~(d).
\vspace{-1em}
}
\label{fig:method-overview}
\end{figure*}

Approaches based on active or weakly-supervised learning~\cite{shi2022weakly} enable using sparse or incomplete labels but commonly need user input to guide initial predictions. Very recently, distilling segmentation foundational models has seen interest for the indoor domain~\cite{guo2023sam,yang2023sam3d,huang2023segment3d} but remains in its infancy for outdoor LiDAR data~\cite{liu2024segment}. In this work, we take an alternative path, proposing an automatic learning-based algorithm for instance-based segmentation of dense aggregated 3D LiDAR point clouds, which requires no semantic or instance annotations during training and produces class-agnostic segmentations that can be mapped to various class nomenclatures and instance granularities for downstream tasks. We make two key design decisions leading to important implications for our method. First, instead of processing individual scans as done by most existing approaches~\cite{nunes2022unsupervised,marcuzzi2023mask}, we follow the manual labeling approach~\cite{behley2019semantickitti} and construct dense registered 3D LiDAR acquisitions to improve the quality of our instance segmentations. By improving the sampling density, this allows us to reliably segment smaller objects and exploit dense image-to-point correspondences, enabling effective use of image-based features. As our second design choice, drawing inspiration from the 3D indoor domain~\cite{wang2023cut,rozenberszki2023unscene3d,zhang2023freepoint}, we design our framework to operate in two stages: \textit{unsupervised generation and self-trained refinement of mask proposals.} 
For generation of our instance mask proposals, we integrate recent deep self-supervised point-\cite{nunes2023temporal} and image-based~\cite{oquab2023dinov2,kirillov2023segany} representations constructed from aligned LiDAR and image captures, providing dense 3D features with high predictive power. We combine these 3D descriptors with spatial priors in a weighted proxy-graph and perform normalized cuts~\cite{shi2000normalized} obtaining initial, possibly noisy, instance segmentations. To eliminate spurious or noisy instances and maximize segmentation performance, we refine mask proposals by performing self-training using a procedure similar to~\cite{rozenberszki2023unscene3d,zhang2023freepoint}. We evaluate our algorithm on SemanticKITTI~\cite{behley2019semantickitti}, mapping our segmentations to ground-truth instances. To quantify advantages of our learning-based algorithm, we compare it to non-learnable~\cite{mcinnes2017hdbscan,yang2023sam3d}, and learnable unsupervised~\cite{nunes2022unsupervised}, as well as supervised~\cite{marcuzzi2023mask}~\cite{yang2023sam3d} instance segmentation baselines. We assess contributions of our components and identify an optimal configuration of features for learning-based instance generation and refinement. 
We contribute:
\begin{itemize}
\item A learning-based algorithm for unsupervised instance-based segmentation of outdoor LiDAR 3D point clouds. 
\item We compare multiple options to build an optimal configuration of self-supervised features for label-free generation and refinement of instance masks, achieving state-of-the-art results against multiple baselines.
\end{itemize}
\section{Related Work}
\label{sec:related}

\paragraph{Panoptic and Instance 3D Segmentation} 
Panoptic~\cite{behley2021benchmark} and instance~\cite{zhang2020instance} 3D LiDAR segmentation are 
key 3D computer vision tasks with vast applications in autonomous robotics, where they serve to partition an input point cloud into distinct object instances and background regions.
Earlier approaches produce instances by spatially clustering~\cite{zhao2022divide,mcinnes2017hdbscan} segmentations predicted by pretrained scene understanding models but depend on the performance of the backbone~\cite{choy20194d,zhu2021cylindrical}.
For realtime operation of mobile robots, learning-based methods have been proposed for panoptic~\cite{marcuzzi2023mask} and instance-based~\cite{zhang2020instance} segmentation of single-scan 3D acquisitions.
Alternatively, segmentation of dense 3D point clouds (similar to this work) can aid annotation of 3D data or perform scene analytics. Here, static indoor scenes have seen the most progress with the emergence of panoptic~\cite{wu2021scenegraphfusion,kolodiazhnyi2023oneformer3d} and instance~\cite{schult2023mask3d} understanding schemes; however instance-based segmentation of dense outdoor LiDAR 3D data remains almost unexplored~\cite{yang2021tupper}. All these approaches are supervised and require large-scale annotated 3D LiDAR datasets~\cite{behley2019semantickitti,fong2022panoptic}.We adopt a recent architectural solution~\cite{marcuzzi2023mask} in our proposal refinement network.

\paragraph{Weakly-, Semi-, and Self-Supervised 3D Segmentation} 
Multiple works construct and utilize sparse annotations (\eg, a single click or a scribble per instance)~\cite{shi2022weakly}, or perform self-supervised pre-training of representations coupled with supervised fine-tuning for respective tasks~\cite{nunes2023temporal,nunes2022segcontrast}. 
Still, they predominantly target either only semantic segmentation~\cite{nunes2022segcontrast,shi2022weakly} or only indoor scenes~\cite{chibane2022box2mask}.
A recent domain adaptation method~\cite{bevsic2022unsupervised} transfers a pre-trained panoptic model between source and target domains, minimizing the gap between them. None of these methods can handle new object types emerging during the inference stage. Weakly- and semi-supervised methods can to some extent reduce the requirements for annotations but are unable to exclude them entirely. We integrate self-supervised point embeddings~\cite{nunes2023temporal} in our instance proposal stage.

\paragraph{Unsupervised 3D Segmentation}
To avoid the need for annotated 3D data and enable integrating object types unseen during training, multiple approaches opt to generate and refine 3D instance proposals without associated semantic labels (\ie, \textit{class-agnostic).} Many such approaches originate from clustering algorithms and use various point cloud representations including raw 3D points~\cite{mcinnes2017hdbscan}, radial projections, range images~\cite{zermas2017fast} and voxel grids~\cite{behley2013laser}. 
Recently, unsupervised pointwise features~\cite{nunes2022segcontrast} were explored as a starting representation for clustering~\cite{nunes2022unsupervised}. 
To ease the segmentation of above ground objects, recent methods leverage ground segmentation~\cite{zermas2017fast, nunes2022unsupervised}; likewise, our approach integrates a robust ground estimator~\cite{lee2022patchwork++} for this purpose. For the indoor domain, recent methods~\cite{rozenberszki2023unscene3d,zhang2023freepoint}  generate 3D instance proposals using deep self-supervised and traditional features, refining proposed masks by a self-trained neural network~\cite{schult2023mask3d}; our work is conceptually similar but targets outdoor 3D LiDAR data. 
Multiple recent approaches~\cite{yang2023sam3d, guo2023sam, huang2023segment3d} distill labels from 2D segmentation foundational models~\cite{kirillov2023segany} into 3D space. 
Among these, we compare to a simple mask merging baseline~\cite{yang2023sam3d}.

\section{Method}
\label{sec:method}

Our algorithm accepts a set of posed LiDAR scans and RGB images as input, and produces dense class-agnostic 3D instance segmentation as output, requiring no manual annotations during training. To this end, we design a learning-based, unsupervised 3D instance segmentation framework (for overview, see \cref{fig:method-overview}). Our key idea is to adopt pre-trained neural representations to model pairwise point similarity, producing (potentially, noisy) 3D instance masks (\cref{method:instance-proposal}) that are refined by a self-training iteration (\cref{method:self-training}). 

\subsection{Scan Preprocessing and Chunk Extraction}
\label{method:preprocessing}

We assume that a set of LiDAR 3D scans~$\{P_t\}_{t=1}^T$ with their world 3D poses~$\{T_t^{\text{P}}\}_{t=1}^T$, as well as a set of images~$\{I_t\}_{t=1}^T$ and their world 3D poses $\{T_t^{\text{I}}\}_{t=1}^T$ are produced by a scanning system including one or more LiDAR sensors and RGB cameras. 
Unless provided, accurate 3D poses can be obtained using GNSS reference poses and further refined with SLAM~\cite{behley2018rss}, due to the synchronization of sensors, adjacent LiDAR scans and image captures can be obtained for each 3D pose in the global map. We further obtain the relevant camera features by projecting the pixels onto the aggregated pointcloud and leverage hidden-point removal.
Operating with individual 3D LiDAR scans~\cite{nunes2022unsupervised} complicates scene understanding tasks as such data rarely has high sampling density~\cite{li2021durlar}; instead, to enable reliable 3D instance segmentation, we compute a dense 3D point cloud. For this, we transform 3D LiDAR and image data to a shared reference frame; 
dynamic objects can be detected and eliminated by recent effective methods~\cite{chen2021moving}. During this aggregation step, we perform ground segmentation~\cite{lee2022patchwork++}, effectively removing points that do not contain instances and allowing to exploit the geometric separation between objects. To reduce memory constraints, we additionally perform voxel downsampling with a voxel size of 5\,cm. Due to the high density of the resulting pointcloud, our method as well the baselines operate on a further down-sampled pointcloud, with predictions being reprojected onto the dense map with 5\,cm voxel size for evaluation.
To enable using image-based features, we build an inverted index associating each point~$\bm{p}$ to a set of pairs~$A_{\bm{p}} = \{(I_{k}, \bm{u}_k)\}$ where $\bm{u}_k$ is a pixel in view~$I_{k}$ capturing~$\bm{p}$. To prevent points occluded in a given view from being assigned its pixels, we identify occluded points \wrt each view~\cite{katz2007direct}.
For our large-scale scenes, processing entire LiDAR 3D point maps is computationally infeasible as they include billions of sampled points; instead, we extract local \textit{overlapping point chunks} for all subsequent steps. Our chunks are $w \times h \times d$ axis-aligned rectangular parallelepipeds, sampled along the cars trajectory to maximize map coverage.
In our experiments, we evaluate the influence of chunk size on performance.

\subsection{Generating 3D Instance Mask Proposals}
\label{method:instance-proposal}

\begin{figure}[h!]
\centerline{
\includegraphics[
width=\columnwidth]{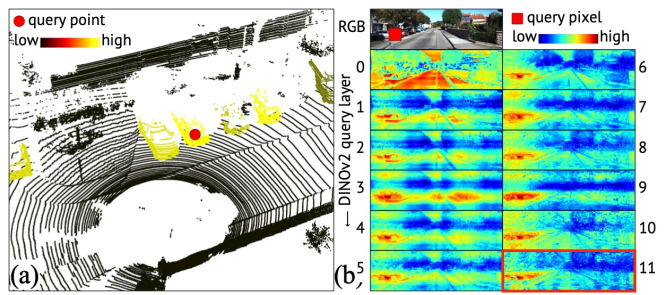}
}
\caption{Pointwise~(a) and pixelwise~(b) similarity maps~\eqref{eq:similarity_score} for TARL~\cite{nunes2023temporal} and DINOv2~\cite{oquab2023dinov2} models, respectively.
Following the intuition from prior research~\cite{keetha2023anyloc}, we select the output of \textit{query-11} (red box) as our image-based feature map.}
\label{fig:dino_layers_choice}
\end{figure}

To overcome the lack of annotated training data, we adopt an unsupervised grouping approach for generation of 3D instance mask proposals. 
Specifically, we link neighbouring 3D points into a weighted proxy-graph and cut this graph, isolating groups of points into instances. 
To effectively model the likelihood of two linked points to belong to the same instance, we weight the graph according to a similarity function integrating multi-modal pointwise representations.


\paragraph{Multi-Modal Pointwise Features}
To fully leverage 3D LiDAR and image data, we represent each point by a collection of features computed from respective data modalities, specifically, spatial point coordinates (S), point embeddings (P), and image embeddings (I). 
Building on the availability of pre-trained neural feature extractors, we adopt state-of-the-art self-supervised TARL~\cite{nunes2023temporal} and DINOv2~\cite{oquab2023dinov2} representations to serve as our point- and image-based features, respectively, as shown in~\cref{fig:dino_layers_choice}. 
TARL~\cite{nunes2023temporal} learns temporally-consistent dense representations of 3D LiDAR scans by pulling together features computed in spatially close points. 
DINOv2~\cite{oquab2023dinov2} is a self-distillation training approach providing dense universal features at high spatial resolution (\eg, $4\times$ higher compared to a recent contrastive method~\cite{wang2021dense}), enabling fine-grained binding to points.
Point-based features~$\mathbf{x}^{\text{P}}_{\bm{p}}$ are pre-computed by a pre-trained TARL model~$\varphi^{\text{P}}$ on single LiDAR 3D scans directly.
To compute an image-based feature~$\mathbf{x}^{\text{I}}_{\bm{p}}$ at point~$\bm{p}$, we use a pre-trained DINOv2 model~$\varphi^{\text{I}}$ to extract feature maps~$\varphi^{\text{I}}(I)$ at layer \textit{query-11,} that was identified, similarly to~\cite{keetha2023anyloc}, by qualitatively analysing the capacity of feature similarity maps to capture instances well (\cref{fig:dino_layers_choice}). 
We compute an average
\begin{equation}
\mathbf{x}^{\text{I}}_{\bm{p}} = \frac{1}{|A_{\bm{p}}|} \sum\limits_{(I, \bm{u}) \in A_{\bm{p}}} \varphi^{\text{I}}(I)_{\text{fmap}(\bm{u})},
\end{equation}
where $\text{fmap}(\bm{u})$ is a position in the feature map corresponding to the pixel~$\bm{u}$.
We complement neural descriptors~$\mathbf{x}^{\text{P}}_{\bm{p}}, \mathbf{x}^{\text{I}}_{\bm{p}}$ by a simple spatial feature~$\mathbf{x}^{\text{S}}_{\bm{p}} = \bm{p}$ to express that two \textit{spatially proximal} points may be naturally assigned to the same instance.


 
\begin{figure}[t!]
\centerline{
\includegraphics[
width=.95\columnwidth]{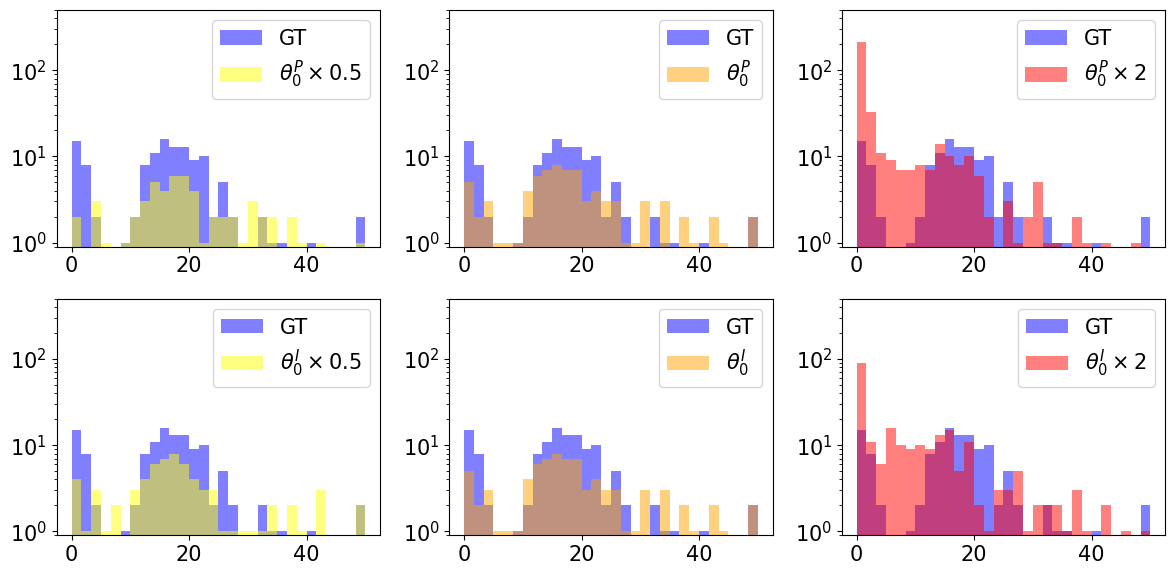}
}
\caption{Influence of hyperparameters $\theta^{\text{P}}, \theta^{\text{I}}$ in~\eqref{eq:similarity_score} on distribution of instance mask proposals, as captured by the volume occupied by instances (horizontal axis). 
$\theta^{\text{P}}_0 = 0.5, \theta^{\text{I}}_0 = 0.1$.}
\label{fig:ablation_weight_thetas_volumes}
\end{figure}

\paragraph{Linking Points into a Proxy-Graph}
To generate instance mask proposals, we start with connecting pairs of points located within a distance~$R = \text{1\,m}$ with weighted edges; we aim to give larger weights to edges within instances and smaller weights to edges crossing instance boundaries.  
For this, we use features constructed in respective spaces to compute pairwise similarity scores 
\begin{equation}
\label{eq:similarity_score}
\mathrm{w}^{\mu}_{\bm{p}\bm{q}} = 
e^{-\theta^{\mu} ||\mathbf{x}^{\mu}_{\bm{p}} - \mathbf{x}^{\mu}_{\bm{q}}||^2},
\end{equation}
where $\mu$ denotes a respective feature space and $\theta^{\mu} > 0$ a scalar parameter. 
We opt to integrate these quantities into a single score using a multiplicative rule
\begin{equation}
\label{eq:integration_of_weights}
\mathrm{w}_{\bm{p}\bm{q}} = \prod\limits_{\mu} \mathrm{w}_{\bm{p}\bm{q}}^{\mu},
\end{equation}
requiring high similariry scores across all modalities for two points to be included in an instance.

\paragraph{Instance Isolation by Normalized Cuts}
\label{method:ncut}
To generate instance mask proposals in a point chunk, we partition the proxy-graph described above using the Normalized Cut (NCut) algorithm~\cite{shi2000normalized}. 
NCut relies on partitioning criteria including both total inter-instance dissimilarity and total intra-instance similarity, naturally matching our task.
At each grouping iteration, we check against the cutting eigenvalue threshold (we use .005 for S+P+I, .03 for S+P, and 0.075 for S) and the minimum share of points in the isolated region (we use 1\%) to prevent further cuts. 
For dense similarity matrices $\mathrm{w}_{\bm{p}\bm{q}}$, singular value decomposition (SVD) employed by NCut can be computationally expensive; by avoiding edges between points located outside radius~$R$ to each other, we obtain a sparser initial matrix. To additionally accommodate for the computational complexity of the Normalized Cut algorithm, we perform uniform voxel downsampling of chunk point (with voxel size 0.35).

\paragraph{Map-Level Instance Segmentation}
To predict a map-level instance segmentation, we merge predictions of individual chunks to the global map by exploiting the overlap between adjacent chunks. For each instance in a new chunk, we determine the instance in the global map with the highest bouding box IoU. Consequently, overlapping instances are merged when the IoU is above a threshold of 0.01.


\subsection{Self-Training of Refined 3D Instance Masks}
\label{method:self-training}

\begin{figure*}[th!]

\centering
\includegraphics[width=\textwidth]{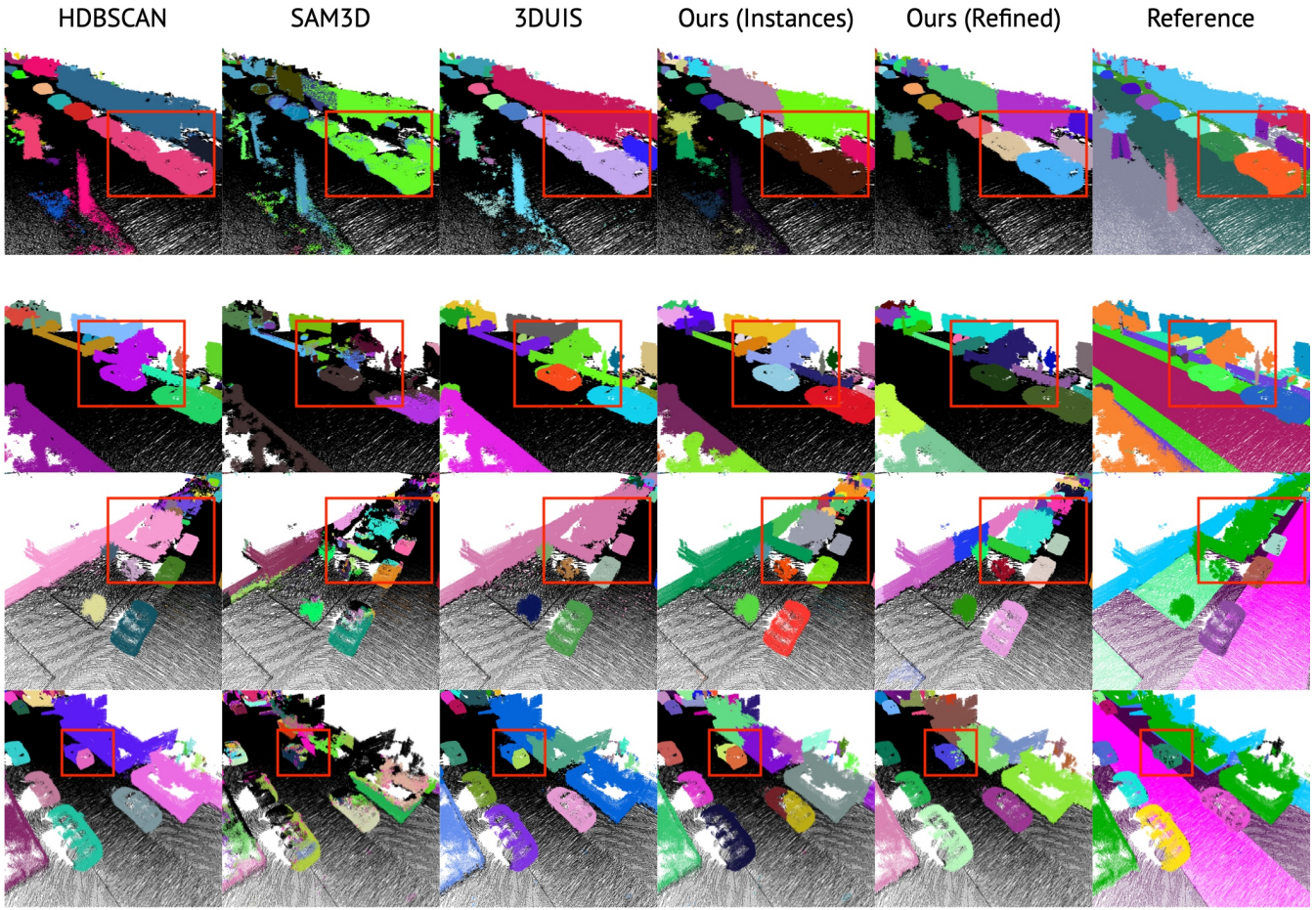}
\caption{\textit{Comparative instance-based segmentation results on SemanticKITTI~\cite{behley2019semantickitti}.}
\vspace{-1em}
}

\label{fig:qual-comp}
\end{figure*}

\begin{table*}[th!]

\centerline{
\includegraphics[width=\textwidth]{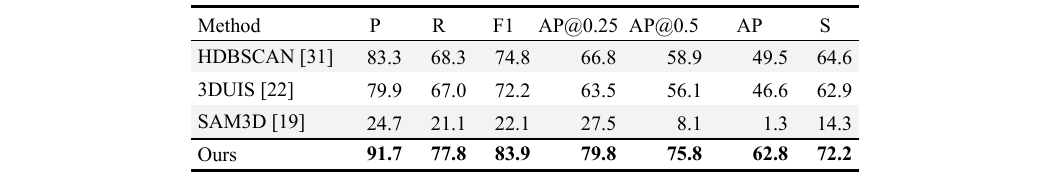}%
}
\caption{\textit{Comparative instance-based segmentation results on SemanticKITTI~\cite{behley2019semantickitti}.}
All measures are in percentages where higher values correspond to better results. 
Our method outperforms all baselines across all quantitative measures.
\vspace{-1em}
}
\label{tbl:comparative}

\end{table*}

\begin{table}[t!]

\centerline{
\includegraphics[
width=\columnwidth,
trim=0 2.4em 0 2.4em,
clip=True
]{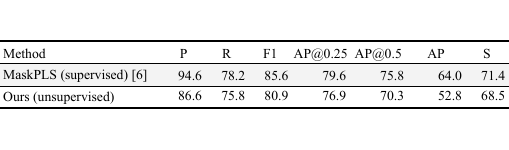}%
}
\caption{\textit{Comparison with the supervised MaskPLS~\cite{marcuzzi2023mask}.}
\vspace{-1em}
}
\label{tbl:comparative_vs_maskpls}

\end{table}

3D masks obtained in~\cref{method:instance-proposal} already provide an instance-based scene description but may include noisy, under- or oversegmented detections, see~\cref{fig:method-overview}.
Instead of treating them as final instances, we take inspiration from recent 3D indoor approaches~\cite{rozenberszki2023unscene3d,zhang2023freepoint,zhang2023towards} and seek to refine these initial predictions by a self-trained neural network (NN).
Intuitively, such approaches (including ours) build on recent studies~\cite{arpit2017closer,ye2021learning} revealing that during training, NNs learn cleaner, simpler patterns first, fitting more noisy data samples later. Drawing on these insights, we design a self-training step that learns a clean chunk-based 3D instance segmentation, starting from the initial set of noisy predictions produced by NCut. Our self-trained NN takes $xyz$ point coordinates as input and produces per-point 3D instance masks and confidence values as output.
To train it, we use the dataset obtained by applying NCut to the collection of point chunks, treating noisy 3D instance masks as ground-truth labels (in practice, we exclude instances represented by fewer than 100~points).
Architecturally, our self-trained NN follows the design of MaskPLS network for  panoptic segmentation~\cite{marcuzzi2023mask}, including a sparse CNN encoder~\cite{choy20194d} and a transformer decoder~\cite{cheng2021per}. 
During training, we optimize a weighted combination of dice and cross-entropy losses
\begin{equation}
\label{eq:self-training-loss}
L =\lambda_{\text{dice}} L_{\text{dice}} + \lambda_{\text{bce}} L_{\text{bce}}.
\end{equation}
where $\lambda_{\text{dice}} = 5$ and $\lambda_{\text{bce}} = 2$ as in~\cite{marcuzzi2023mask}.
We train our network for one iteration, \ie until convergence with initial ground-truth labels; we find this to be sufficient for achieving improved instance segmentation performance. 
Due to the significant density of our aggregated chunk-based point clouds, we preprocess our chunks using uniform downsampling.

\section{Experiments}
\label{sec:experiments}

\subsection{Experimental Setup}
\label{experiments:setup}

\begin{figure*}[th!]

\centering
\includegraphics[width=\textwidth]{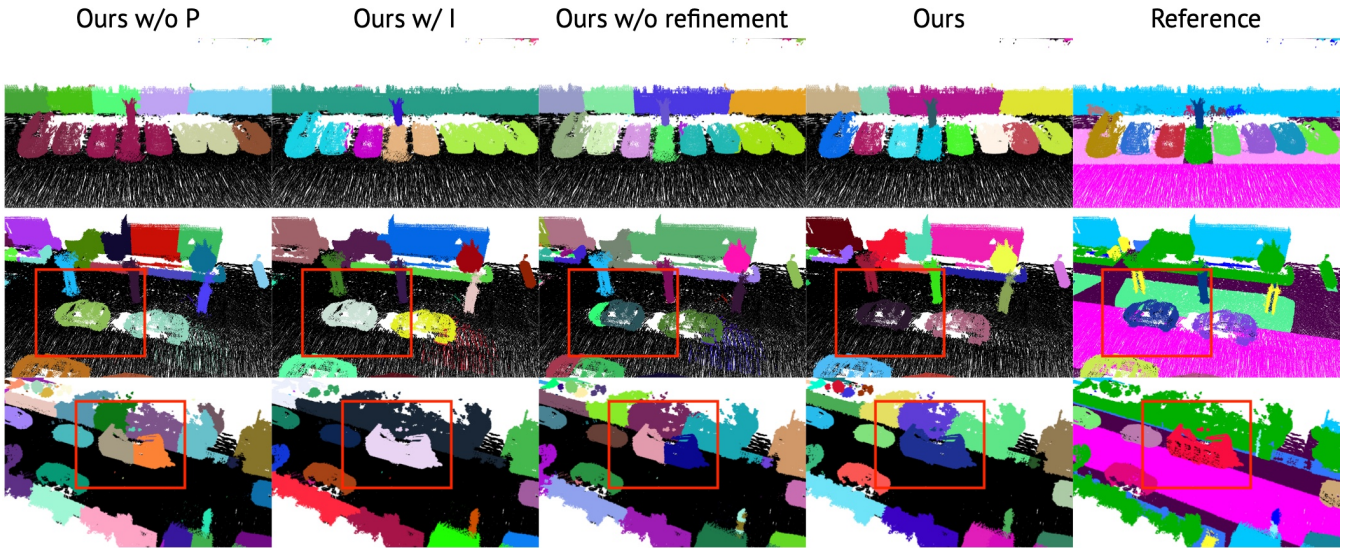}
\caption{\textit{Ablative instance-based segmentation results on SemanticKITTI~\cite{behley2019semantickitti}.}
\vspace{-1em}
}

\label{fig:ablative-comp}
\end{figure*}

\begin{table*}[ht!]

\centerline{
\includegraphics[width=\textwidth]{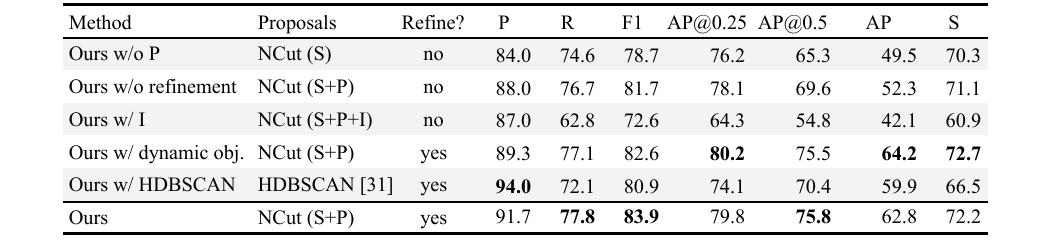}}
\caption{\textit{Quantitative comparison of contribution of various components of our method on SemanticKITTI~\cite{behley2019semantickitti}.} 
All measures are in percentages where higher values correspond to better results.
\enquote{Ours w/o P}, \enquote{Ours w/o refinement}, and \enquote{Ours w/ I} refer to forming the proxy-graph by weights computed from spatial only, spatial and point, and spatial, point, and image features (Proposals column).
\enquote{Ours w/ dynamic obj.} performs segmentation without prior removal of dynamic instances.
\enquote{Ours w/ HDBSCAN} replaces our instance generation step by HDBSCAN~\cite{mcinnes2017hdbscan}.
\vspace{-1em}
}
\label{tbl:ablative}

\end{table*}

\paragraph{Data Preprocessing and Training Details}
We experimentally evaluate our algorithm using SemanticKITTI~\cite{behley2019semantickitti} which provides LiDAR 3D scans with instance annotations and registered RGB images acquired by a moving car in outdoor scenes, obtaining scenes on the order of $10^9$ points after aggregation.
We use sequences 0--10 providing test labels, except sequences~1 and~4 that predominantly consist of dynamic instances. 
For accurate global alignment of LiDAR 3D scans, we use 3D poses computed using a surfel-based SLAM approach~\cite{behley2018rss}. 
During mask proposal and refinement stages, we operate on 25$\times$25$\times$25\,m$^3$ point chunks; at inference time, we sample chunk centers along vehicle trajectory spaced 22\,m apart (3\,m overlap), resulting in 30~chunks per 1K~raw LiDAR scans on average. We extract image-based features from the images obtained by the left camera. 
For training our refinement network and the supervised baseline, we sample point chunks, our proposals, and ground-truth labels, from KITTI sequences 0--10 with a larger 24\,m overlap, obtaining a dataset with 14K chunks. To reduce memory consumption, we uniformly subsample each chunk to roughly $60$K points. We train our refinement NN~\cite{marcuzzi2023mask} for 7 epochs ($12$~h) on RTX 4090 GPU using Adam~\cite{kingma2014adam} with a learning rate of $10^{-4}$. 

\paragraph{Implementation Details}
We set $\theta^{\text{I}}=0.1$, $\theta^{\text{P}}=0.5$ for S+P+I. For S+P, the point feature weight is set to $\theta^{\text{P}}=0.5$, all of our configurations use a spatial weight of $\theta^{\text{S}}=1$. We use the cutting thresholds as mentioned in~\cref{method:instance-proposal}.
To generate the final instance masks, we run inference on downsampled input chunks and transfer the predicted labels to the denser point cloud by nearest-neighbor assignment. 
We study the impact of parameter choices in~\cref{experiments:ablative}.

\paragraph{Metrics} 
Our method is class-agnostic; thus, we evaluate it using instance-based annotations and report instance-related metrics, averaging them across available sequences. 
Specifically, we use Precision (P), Recall (R), and their harmonic mean, F1-score (F1), Average Precision (AP, computed at three IoU thresholds of 0.25, 0.5, and globally), and $S_{\text{assoc}}$~\cite{nunes2022unsupervised} (S).



\paragraph{Baselines} 
We define two unsupervised and two supervised baselines.
HDBSCAN~\cite{mcinnes2017hdbscan} is a density-based clustering algorithm, while
3DUIS~\cite{nunes2022unsupervised} is an unsupervised instance segmentation method, refining HDBSCAN instance proposals with deep point-based features~\cite{nunes2022segcontrast} and graph cuts.
SAM3D~\cite{yang2023sam3d} is a method merging 2D SAM~\cite{guo2023sam} predictions into 3D instance masks.
We pre-process 2D SAM masks by rejecting smaller sub-instances using an IoU threshold of 0.9 and perform hidden point removal. Additionally, we train supervised MaskPLS~\cite{marcuzzi2023mask}, a state-of-the-art 3D panoptic segmentation method. For experiments in Tab. \ref{tbl:comparative_vs_maskpls}, we train on sequences 0--7 and 9, holding out 8 and 10 as the test set. Further experiments (see Tab. \ref{tbl:comparative} and Tab. \ref{tbl:ablative}), were tested on all sequences and do not include supervised training.

\subsection{Comparisons to State-of-the-Art}
\label{experiments:comparative}

We start by evaluating our method against baselines; we present quantitative evaluation results against the unsupervised baselines in \cref{tbl:comparative} and a quantitative comparison against the supervised MaskPLS method~\cite{marcuzzi2023mask} in \cref{tbl:comparative_vs_maskpls}.
Our best-performing method (NCut(S + P) + Refine) outperforms point-based methods HDBSCAN and 3DUIS and the supervised method SAM3D without using image-based features. We refer to our ablations~\ref{experiments:ablative} for a comparison of our point-based method (Ncuts(S+P)) with a multi-modal implementation (NCut(S + P + I)). While our approach shows a moderate decrease \wrt the fully supervised MaskPLS, it does not require manual labeling and can be used to speed up the annotation for supervised training scenarios. Qualitative comparisons are presented in~\cref{fig:qual-comp} for baselines and our algorithm without and with refinement step. One noticeable effect of refinement is that it splits instances that are merged together in initial proposals.

\begin{table}[t]

\centerline{
\includegraphics[width=\columnwidth]{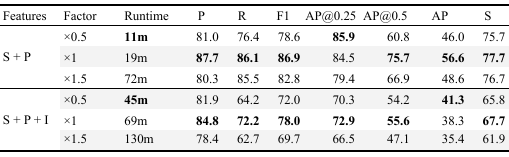}}
\caption{\textit{Influence of chunk size on performance.}
Our chunks provide sufficient context for effective proposal generation at moderate computational cost.
\vspace{-1em}
}
\label{tbl:ablation_chunk_size}

\end{table}

\begin{table}[t]

\centerline{
\includegraphics[width=\columnwidth]{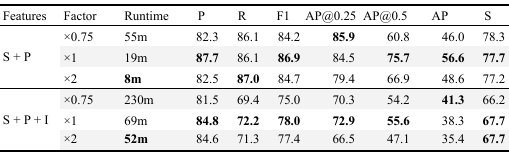}}
\caption{\textit{Influence of voxel size on performance.}
The base voxel size of 35\,cm provides a good balance between computational cost and performance.
\vspace{-1em}
}
\label{tbl:ablation_voxel_size}

\end{table}

\subsection{Ablative Studies}
\label{experiments:ablative}

\paragraph{Making Effective Instance Proposals}
To identify the optimal ingredients in our instance mask proposal step, we compare applying NCut to point graphs weighted by various combination of features, without any refinement, and present the results in~\cref{tbl:ablative} (see notation in its caption).
We conclude that point embeddings (P) are crucial for obtaining strong performance; in contrast (and to our surprise), image embeddings (I) noticeably decrease performance across measures. To understand whether an alternative proposal generation step may lead to obtaining comparably strong results, we generate instance mask proposals by HDBSCAN and refine them using our self-training step. While this improves upon HDBSCAN, our method achieves better results.

\paragraph{Fine-Grained Tuning of Proposal Hyperparameters}
We assess the influence of chunk and voxel size used by evaluating the performance and runtime of our method, these ablations are performed on a single sequence of KITTI~\cite{behley2019semantickitti}. Results presented in \cref{tbl:ablation_chunk_size,tbl:ablation_voxel_size} indicate that, for our configuration and dataset, our default chunk size (25$\times$25$\times$25\,m$^3$) and voxel size (0.35\,m) are close to optimal. Larger voxel size leads to coarser instances while finer one yields noisier instances.
We additionally made an effort to find an optimal combination of parameters $\theta^{\text{S}}, \theta^{\text{P}}, \theta^{\text{I}}$ in~\eqref{eq:similarity_score}. From \cref{fig:ablation_weight_thetas_volumes} we conclude that our default values lead our proposals to match the organic instance distributions.

\begin{table}[t!]

\centerline{
\includegraphics[
width=\columnwidth,
trim=0 1em 0 1em,
clip=True,
]{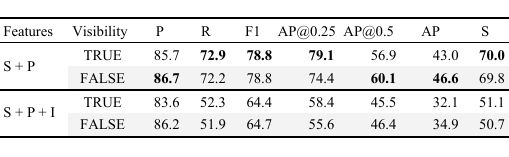}}
\caption{\textit{Influence of point visibility.}
Detecting and eliminating invisible points moderately affects performance.
\vspace{-1em}}
\label{tbl:ablation_visibility}

\end{table}

\begin{figure}[t!]
\centerline{
\includegraphics[
width=\columnwidth]{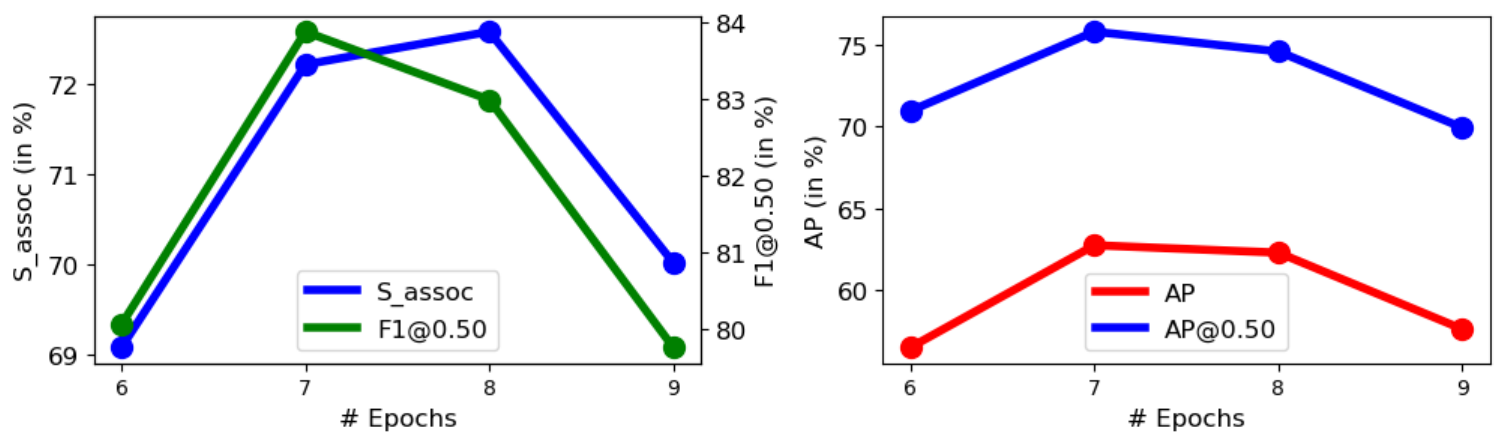}
}
\caption{\textit{Influence of self-training duration on performance.}
\vspace{-1em}}
\label{fig:num-self-training-epochs}
\end{figure}

\paragraph{The Contribution of Refinement} Convergence of self-supervised refinement for KITTI dataset is depicted in \cref{fig:num-self-training-epochs}. As demonstrated from the AP, F1 and $S_{assoc}$ results, 7 epochs is enough for the network to be saturated.

\paragraph{Robustness \wrt Dynamic Objects}
In \cref{tbl:ablative}, we include an additional result to illustrate robustness of our method \wrt elimination of dynamic instances (\enquote{Ours w/ dynamic obj.}).
We aggregate 3D scans without filtering dynamic objects; we then perform instance generation and refinement as usual. 
We find that training on such noisier data yields almost no drop in performance (\eg, only 1.3\% in F1).

\paragraph{Influence of Visibility on Image Embeddings} 
We assess whether eliminating points predicted as hidden in all images from the joint point cloud enhances performance. We employ hidden point removal (radius=20) to eliminate invisible points, applying NCut on visible point only. \cref{tbl:ablation_visibility} suggests that visibility does not have a significant impact on performance, regardless of the feature configuration. 

\section{Conclusion}
\label{sec:conclusion}

We have developed a learning-based, label-free algorithm for instance-based segmentation of outdoor LiDAR 3D point clouds. 
Our method uses dense, registered 3D and RGB data to perform generation of dense 3D instance mask proposals, constructing and cutting a feature-aware graph. 
We have discovered that integrating self-supervised point embeddings with simple spatial features results in the best performance across feature combinations. 
Our algorithm further benefits from a self-training step that does not require ground-truth annotations, instead fitting and refining proposals obtained in an unsupervised way.
Our results suggest a better performance compared to clustering-based and learning-based baselines across a variety of instance-sensitive measures.

{\small
\bibliographystyle{IEEEtran}
\bibliography{IEEEabrv,references}

\begin{thebibliography}{10}
\providecommand{\url}[1]{#1}
\csname url@rmstyle\endcsname
\providecommand{\newblock}{\relax}
\providecommand{\bibinfo}[2]{#2}
\providecommand\BIBentrySTDinterwordspacing{\spaceskip=0pt\relax}
\providecommand\BIBentryALTinterwordstretchfactor{4}
\providecommand\BIBentryALTinterwordspacing{\spaceskip=\fontdimen2\font plus
\BIBentryALTinterwordstretchfactor\fontdimen3\font minus
  \fontdimen4\font\relax}
\providecommand\BIBforeignlanguage[2]{{%
\expandafter\ifx\csname l@#1\endcsname\relax
\typeout{** WARNING: IEEEtran.bst: No hyphenation pattern has been}%
\typeout{** loaded for the language `#1'. Using the pattern for}%
\typeout{** the default language instead.}%
\else
\language=\csname l@#1\endcsname
\fi
#2}}

\bibitem{marcuzzi2023mask}
R.~Marcuzzi, L.~Nunes, L.~Wiesmann, J.~Behley, and C.~Stachniss, ``Mask-based
  panoptic lidar segmentation for autonomous driving,'' \emph{IEEE Robotics and
  Automation Letters}, vol.~8, no.~2, pp. 1141--1148, 2023.

\bibitem{li2023cpseg}
E.~Li, R.~Razani, Y.~Xu, and B.~Liu, ``Cpseg: Cluster-free panoptic
  segmentation of 3d lidar point clouds,'' in \emph{2023 IEEE International
  Conference on Robotics and Automation (ICRA)}.\hskip 1em plus 0.5em minus
  0.4em\relax IEEE, 2023, pp. 8239--8245.

\bibitem{behley2019semantickitti}
J.~Behley, M.~Garbade, A.~Milioto, J.~Quenzel, S.~Behnke, C.~Stachniss, and
  J.~Gall, ``Semantickitti: A dataset for semantic scene understanding of lidar
  sequences,'' in \emph{Proceedings of the IEEE/CVF international conference on
  computer vision}, 2019, pp. 9297--9307.

\bibitem{dai2017scannet}
A.~Dai, A.~X. Chang, M.~Savva, M.~Halber, T.~Funkhouser, and M.~Nie{\ss}ner,
  ``Scannet: Richly-annotated 3d reconstructions of indoor scenes,'' in
  \emph{Proc. Computer Vision and Pattern Recognition (CVPR), IEEE}, 2017.

\bibitem{yeshwanthliu2023scannetpp}
C.~Yeshwanth, Y.-C. Liu, M.~Nie{\ss}ner, and A.~Dai, ``Scannet++: A
  high-fidelity dataset of 3d indoor scenes,'' in \emph{Proceedings of the
  International Conference on Computer Vision ({ICCV})}, 2023.

\bibitem{huang2023segment3d}
R.~Huang, S.~Peng, A.~Takmaz, F.~Tombari, M.~Pollefeys, S.~Song, G.~Huang, and
  F.~Engelmann, ``Segment3d: Learning fine-grained class-agnostic 3d
  segmentation without manual labels,'' \emph{arXiv preprint arXiv:2312.17232},
  2023.

\bibitem{bevsic2022unsupervised}
B.~Be{\v{s}}i{\'c}, N.~Gosala, D.~Cattaneo, and A.~Valada, ``Unsupervised
  domain adaptation for lidar panoptic segmentation,'' \emph{IEEE Robotics and
  Automation Letters}, vol.~7, no.~2, pp. 3404--3411, 2022.

\bibitem{shi2022weakly}
H.~Shi, J.~Wei, R.~Li, F.~Liu, and G.~Lin, ``Weakly supervised segmentation on
  outdoor 4d point clouds with temporal matching and spatial graph
  propagation,'' in \emph{Proceedings of the IEEE/CVF Conference on Computer
  Vision and Pattern Recognition}, 2022, pp. 11\,840--11\,849.

\bibitem{guo2023sam}
H.~Guo, H.~Zhu, S.~Peng, Y.~Wang, Y.~Shen, R.~Hu, and X.~Zhou, ``Sam-guided
  graph cut for 3d instance segmentation,'' \emph{arXiv preprint
  arXiv:2312.08372}, 2023.

\bibitem{yang2023sam3d}
Y.~Yang, X.~Wu, T.~He, H.~Zhao, and X.~Liu, ``Sam3d: Segment anything in 3d
  scenes,'' \emph{arXiv preprint arXiv:2306.03908}, 2023.

\bibitem{liu2024segment}
Y.~Liu, L.~Kong, J.~Cen, R.~Chen, W.~Zhang, L.~Pan, K.~Chen, and Z.~Liu,
  ``Segment any point cloud sequences by distilling vision foundation models,''
  \emph{Advances in Neural Information Processing Systems}, vol.~36, 2024.

\bibitem{nunes2022unsupervised}
L.~Nunes, X.~Chen, R.~Marcuzzi, A.~Osep, L.~Leal-Taix{\'e}, C.~Stachniss, and
  J.~Behley, ``Unsupervised class-agnostic instance segmentation of 3d lidar
  data for autonomous vehicles,'' \emph{IEEE Robotics and Automation Letters},
  vol.~7, no.~4, pp. 8713--8720, 2022.

\bibitem{wang2023cut}
X.~Wang, R.~Girdhar, S.~X. Yu, and I.~Misra, ``Cut and learn for unsupervised
  object detection and instance segmentation,'' in \emph{Proceedings of the
  IEEE/CVF Conference on Computer Vision and Pattern Recognition}, 2023, pp.
  3124--3134.

\bibitem{rozenberszki2023unscene3d}
D.~Rozenberszki, O.~Litany, and A.~Dai, ``Unscene3d: Unsupervised 3d instance
  segmentation for indoor scenes,'' \emph{arXiv preprint arXiv:2303.14541},
  2023.

\bibitem{zhang2023freepoint}
Z.~Zhang, J.~Ding, L.~Jiang, D.~Dai, and G.-S. Xia, ``Freepoint: Unsupervised
  point cloud instance segmentation,'' \emph{arXiv preprint arXiv:2305.06973},
  2023.

\bibitem{nunes2023temporal}
L.~Nunes, L.~Wiesmann, R.~Marcuzzi, X.~Chen, J.~Behley, and C.~Stachniss,
  ``Temporal consistent 3d lidar representation learning for semantic
  perception in autonomous driving,'' in \emph{Proceedings of the IEEE/CVF
  Conference on Computer Vision and Pattern Recognition}, 2023, pp. 5217--5228.

\bibitem{oquab2023dinov2}
M.~Oquab, T.~Darcet, T.~Moutakanni, H.~Vo, M.~Szafraniec, V.~Khalidov,
  P.~Fernandez, D.~Haziza, F.~Massa, A.~El-Nouby, \emph{et~al.}, ``Dinov2:
  Learning robust visual features without supervision,'' \emph{arXiv preprint
  arXiv:2304.07193}, 2023.

\bibitem{kirillov2023segany}
A.~Kirillov, E.~Mintun, N.~Ravi, H.~Mao, C.~Rolland, L.~Gustafson, T.~Xiao,
  S.~Whitehead, A.~C. Berg, W.-Y. Lo, P.~Doll{\'a}r, and R.~Girshick, ``Segment
  anything,'' \emph{arXiv:2304.02643}, 2023.

\bibitem{shi2000normalized}
J.~Shi and J.~Malik, ``Normalized cuts and image segmentation,'' \emph{IEEE
  Transactions on pattern analysis and machine intelligence}, vol.~22, no.~8,
  pp. 888--905, 2000.

\bibitem{mcinnes2017hdbscan}
L.~McInnes, J.~Healy, and S.~Astels, ``hdbscan: Hierarchical density based
  clustering.'' \emph{J. Open Source Softw.}, vol.~2, no.~11, p. 205, 2017.

\bibitem{behley2021benchmark}
J.~Behley, A.~Milioto, and C.~Stachniss, ``A benchmark for lidar-based panoptic
  segmentation based on kitti,'' in \emph{2021 IEEE International Conference on
  Robotics and Automation (ICRA)}.\hskip 1em plus 0.5em minus 0.4em\relax IEEE,
  2021, pp. 13\,596--13\,603.

\bibitem{zhang2020instance}
F.~Zhang, C.~Guan, J.~Fang, S.~Bai, R.~Yang, P.~H. Torr, and V.~Prisacariu,
  ``Instance segmentation of lidar point clouds,'' in \emph{2020 IEEE
  International Conference on Robotics and Automation (ICRA)}.\hskip 1em plus
  0.5em minus 0.4em\relax IEEE, 2020, pp. 9448--9455.

\bibitem{zhao2022divide}
Y.~Zhao, X.~Zhang, and X.~Huang, ``A divide-and-merge point cloud clustering
  algorithm for lidar panoptic segmentation,'' in \emph{2022 International
  Conference on Robotics and Automation (ICRA)}.\hskip 1em plus 0.5em minus
  0.4em\relax IEEE, 2022, pp. 7029--7035.

\bibitem{choy20194d}
C.~Choy, J.~Gwak, and S.~Savarese, ``4d spatio-temporal convnets: Minkowski
  convolutional neural networks,'' in \emph{Proceedings of the IEEE/CVF
  conference on computer vision and pattern recognition}, 2019, pp. 3075--3084.

\bibitem{zhu2021cylindrical}
X.~Zhu, H.~Zhou, T.~Wang, F.~Hong, Y.~Ma, W.~Li, H.~Li, and D.~Lin,
  ``Cylindrical and asymmetrical 3d convolution networks for lidar
  segmentation,'' in \emph{Proceedings of the IEEE/CVF conference on computer
  vision and pattern recognition}, 2021, pp. 9939--9948.

\bibitem{wu2021scenegraphfusion}
S.-C. Wu, J.~Wald, K.~Tateno, N.~Navab, and F.~Tombari, ``Scenegraphfusion:
  Incremental 3d scene graph prediction from rgb-d sequences,'' in
  \emph{Proceedings of the IEEE/CVF Conference on Computer Vision and Pattern
  Recognition}, 2021, pp. 7515--7525.

\bibitem{kolodiazhnyi2023oneformer3d}
M.~Kolodiazhnyi, A.~Vorontsova, A.~Konushin, and D.~Rukhovich, ``Oneformer3d:
  One transformer for unified point cloud segmentation,'' \emph{arXiv preprint
  arXiv:2311.14405}, 2023.

\bibitem{schult2023mask3d}
J.~Schult, F.~Engelmann, A.~Hermans, O.~Litany, S.~Tang, and B.~Leibe,
  ``Mask3d: Mask transformer for 3d semantic instance segmentation,'' in
  \emph{2023 IEEE International Conference on Robotics and Automation
  (ICRA)}.\hskip 1em plus 0.5em minus 0.4em\relax IEEE, 2023, pp. 8216--8223.

\bibitem{yang2021tupper}
Z.~Yang and C.~Liu, ``Tupper-map: Temporal and unified panoptic perception for
  3d metric-semantic mapping,'' in \emph{2021 IEEE/RSJ International Conference
  on Intelligent Robots and Systems (IROS)}.\hskip 1em plus 0.5em minus
  0.4em\relax IEEE, 2021, pp. 1094--1101.

\bibitem{fong2022panoptic}
W.~K. Fong, R.~Mohan, J.~V. Hurtado, L.~Zhou, H.~Caesar, O.~Beijbom, and
  A.~Valada, ``Panoptic nuscenes: A large-scale benchmark for lidar panoptic
  segmentation and tracking,'' \emph{IEEE Robotics and Automation Letters},
  vol.~7, no.~2, pp. 3795--3802, 2022.

\bibitem{nunes2022segcontrast}
L.~Nunes, R.~Marcuzzi, X.~Chen, J.~Behley, and C.~Stachniss, ``Segcontrast: 3d
  point cloud feature representation learning through self-supervised segment
  discrimination,'' \emph{IEEE Robotics and Automation Letters}, vol.~7, no.~2,
  pp. 2116--2123, 2022.

\bibitem{chibane2022box2mask}
J.~Chibane, F.~Engelmann, T.~Anh~Tran, and G.~Pons-Moll, ``Box2mask: Weakly
  supervised 3d semantic instance segmentation using bounding boxes,'' in
  \emph{European Conference on Computer Vision}.\hskip 1em plus 0.5em minus
  0.4em\relax Springer, 2022, pp. 681--699.

\bibitem{zermas2017fast}
D.~Zermas, I.~Izzat, and N.~Papanikolopoulos, ``Fast segmentation of 3d point
  clouds: A paradigm on lidar data for autonomous vehicle applications,'' in
  \emph{2017 IEEE International Conference on Robotics and Automation
  (ICRA)}.\hskip 1em plus 0.5em minus 0.4em\relax IEEE, 2017, pp. 5067--5073.

\bibitem{behley2013laser}
J.~Behley, V.~Steinhage, and A.~B. Cremers, ``Laser-based segment
  classification using a mixture of bag-of-words,'' in \emph{2013 IEEE/RSJ
  International Conference on Intelligent Robots and Systems}.\hskip 1em plus
  0.5em minus 0.4em\relax IEEE, 2013, pp. 4195--4200.

\bibitem{lee2022patchwork++}
S.~Lee, H.~Lim, and H.~Myung, ``Patchwork++: Fast and robust ground
  segmentation solving partial under-segmentation using 3d point cloud,'' in
  \emph{2022 IEEE/RSJ International Conference on Intelligent Robots and
  Systems (IROS)}.\hskip 1em plus 0.5em minus 0.4em\relax IEEE, 2022, pp.
  13\,276--13\,283.

\bibitem{behley2018rss}
J.~Behley and C.~Stachniss, ``Efficient surfel-based slam using 3d laser range
  data in urban environments,'' in \emph{Proc.~of Robotics: Science and
  Systems~(RSS)}, 2018.

\bibitem{li2021durlar}
L.~Li, K.~N. Ismail, H.~P. Shum, and T.~P. Breckon, ``Durlar: A high-fidelity
  128-channel lidar dataset with panoramic ambient and reflectivity imagery for
  multi-modal autonomous driving applications,'' in \emph{2021 International
  Conference on 3D Vision (3DV)}.\hskip 1em plus 0.5em minus 0.4em\relax IEEE,
  2021, pp. 1227--1237.

\bibitem{chen2021moving}
X.~Chen, S.~Li, B.~Mersch, L.~Wiesmann, J.~Gall, J.~Behley, and C.~Stachniss,
  ``Moving object segmentation in 3d lidar data: A learning-based approach
  exploiting sequential data,'' \emph{IEEE Robotics and Automation Letters},
  vol.~6, no.~4, pp. 6529--6536, 2021.

\bibitem{katz2007direct}
S.~Katz, A.~Tal, and R.~Basri, ``Direct visibility of point sets,'' in
  \emph{ACM SIGGRAPH 2007 papers}, 2007, pp. 24--es.

\bibitem{keetha2023anyloc}
N.~Keetha, A.~Mishra, J.~Karhade, K.~M. Jatavallabhula, S.~Scherer, M.~Krishna,
  and S.~Garg, ``Anyloc: Towards universal visual place recognition,''
  \emph{IEEE Robotics and Automation Letters}, 2023.

\bibitem{wang2021dense}
X.~Wang, R.~Zhang, C.~Shen, T.~Kong, and L.~Li, ``Dense contrastive learning
  for self-supervised visual pre-training,'' in \emph{Proceedings of the
  IEEE/CVF conference on computer vision and pattern recognition}, 2021, pp.
  3024--3033.

\bibitem{zhang2023towards}
L.~Zhang, A.~J. Yang, Y.~Xiong, S.~Casas, B.~Yang, M.~Ren, and R.~Urtasun,
  ``Towards unsupervised object detection from lidar point clouds,'' in
  \emph{Proceedings of the IEEE/CVF Conference on Computer Vision and Pattern
  Recognition}, 2023, pp. 9317--9328.

\bibitem{arpit2017closer}
D.~Arpit, S.~Jastrzkebski, N.~Ballas, D.~Krueger, E.~Bengio, M.~S. Kanwal,
  T.~Maharaj, A.~Fischer, A.~Courville, Y.~Bengio, \emph{et~al.}, ``A closer
  look at memorization in deep networks,'' in \emph{International conference on
  machine learning}.\hskip 1em plus 0.5em minus 0.4em\relax PMLR, 2017, pp.
  233--242.

\bibitem{ye2021learning}
S.~Ye, D.~Chen, S.~Han, and J.~Liao, ``Learning with noisy labels for robust
  point cloud segmentation,'' in \emph{Proceedings of the IEEE/CVF
  international conference on computer vision}, 2021, pp. 6443--6452.

\bibitem{cheng2021per}
B.~Cheng, A.~Schwing, and A.~Kirillov, ``Per-pixel classification is not all
  you need for semantic segmentation,'' \emph{Advances in neural information
  processing systems}, vol.~34, pp. 17\,864--17\,875, 2021.

\bibitem{kingma2014adam}
D.~P. Kingma and J.~Ba, ``Adam: A method for stochastic optimization,''
  \emph{arXiv preprint arXiv:1412.6980}, 2014.

\end{thebibliography}
 }

\end{document}